\documentclass[11pt,a4paper]{article}
\usepackage{times,latexsym}
\usepackage{url}
\usepackage[T1]{fontenc}
\usepackage{graphicx}
\usepackage{booktabs}
\usepackage{multirow}

\usepackage{xcolor}
\usepackage{tcolorbox}

\usepackage{tikz}
\newcommand*\circled[1]{\tikz[baseline=(char.base)]{\node[shape=circle,draw,inner sep=2pt] (char) {#1};}}

\definecolor{headerbox}{HTML}{7BAFDE}
\definecolor{textbox}{HTML}{F5F5F5}
\definecolor{cblue}{HTML}{77AADD} 
\definecolor{cmint}{HTML}{44BB99} 
\definecolor{cgreen}{HTML}{AAAA00}
\definecolor{corange}{HTML}{EE8866}

\usepackage[acceptedWithA]{tacl2021v1}
\usepackage{xspace,mfirstuc,tabulary}

\newif\iftaclinstructions
\taclinstructionsfalse %
\iftaclinstructions
\renewcommand{\confidential}{}
\renewcommand{\anonsubtext}{(No author info supplied here, for consistency with
TACL-submission anonymization requirements)}
\newcommand{\instr}
\fi

\iftaclpubformat %

\else

\fi

\title{CRAFT Your Dataset: Task-Specific Synthetic Dataset Generation\\
Through Corpus Retrieval and Augmentation}

\author{
  Ingo Ziegler\textsuperscript{$\diamond$$\star$} \quad
  Abdullatif K\"oksal\textsuperscript{$\dagger$$\ddagger$$\star$} \quad
  Desmond Elliott\textsuperscript{$\diamond$} \quad
  Hinrich Sch\"utze\textsuperscript{$\dagger$$\ddagger$}
  \\
  \ \\
  \textsuperscript{$\diamond$}Department of Computer Science, University of Copenhagen \\
  \textsuperscript{$\dagger$}Center for Information and Language Processing (CIS), LMU Munich \\
  \textsuperscript{$\ddagger$}Munich Center for Machine Learning (MCML) \\
  \textsuperscript{$\star$}Shared first authorship \\
  \href{mailto:inzi@di.ku.dk}{\nolinkurl{inzi@di.ku.dk}}, \href{mailto:akoksal@cis.lmu.de}{\nolinkurl{akoksal@cis.lmu.de}}
}

\date{}

\newcounter{notecounter}
\newcommand{\enotesoff}{\long\gdef\enote##1##2{}}
\newcommand{\enoteson}{\long\gdef\enote##1##2{{
\stepcounter{notecounter}
{\large\bf
\hspace{0cm}\arabic{notecounter} $<<<$ ##1: ##2
$>>>$\hspace{1cm}}}}}

\enoteson
\enotesoff

\long\def\devour#1{\ignorespaces}

\begin{document}
\maketitle
\begin{abstract}
Building high-quality datasets for specialized tasks is a time-consuming and resource-intensive process that often requires specialized domain knowledge.
We propose Corpus Retrieval and Augmentation for Fine-Tuning (CRAFT), a method for generating synthetic datasets, given a small number of user-written few-shots that demonstrate the task to be performed.
Given these examples, CRAFT uses large-scale public web-crawled corpora and similarity-based document retrieval to find other relevant human-written documents.
Lastly, instruction-tuned large language models (LLMs) augment the retrieved documents into custom-formatted task samples, which then can be used for fine-tuning.
We demonstrate that CRAFT can efficiently generate
large-scale task-specific training datasets for four diverse
tasks: biology, medicine, and commonsense question-answering (QA), as well as summarization.
Our experiments show that CRAFT-based models outperform or match general LLMs on QA tasks, while exceeding models trained on human-curated summarization data by 46 preference points.
CRAFT outperforms other synthetic dataset generation methods such as Self- and Evol-Instruct, and remains robust even when the quality of the initial few-shots varies.
\end{abstract}

\section{Introduction}
Large language models (LLMs) demonstrate strong
generalization capabilities across diverse tasks
\cite{dubey2024llama3herdmodels, ouyang_2022}, but optimizing these models for specific tasks remains a
considerable challenge. Although zero-shot and few-shot
prompting methods provide some degree of adaptability
\cite{dong2024surveyincontextlearning}, task-specific
fine-tuning generally delivers better performance,
particularly for specialized and out-of-domain tasks
\cite{liu2022few}. A key challenge for effective fine-tuning
is obtaining high-quality task-specific examples at large scale.

Traditionally, creating high-quality datasets for specific tasks involves a time-consuming and resource-intensive process, often requiring extensive manual curation and annotation (e.g., \citet{marcus-etal-1993-building}). This challenge is particularly acute for low-resource domains or novel tasks where existing datasets may be limited or non-existent.

On the other hand, ``raw'' (i.e., unannotated, free-text) web-crawled corpora are
known for their diversity and potential utility for various
tasks \cite{maini2024rephrasing}. Prior work has used raw data by targeted crawling of recipe websites~\citep{bien-etal-2020-recipenlg} or word-specific filtering of crawling metadata to gather examples from pre-training corpora for sentiment analysis and summarization tasks via ratings~\citep{maas-etal-2011-learning} and bullet point summaries found in news articles~\citep{see-etal-2017-get}.
These approaches either rely on a predefined task definition
based on keywords, or on the targeted crawling of websites
which are expected to contain the desired content.
This reliance hinders the generalization of these methods to tasks where such prior knowledge is unavailable, difficult to define, or highly context-dependent.

In this work, we propose Corpus Retrieval and
Augmentation for Fine-Tuning (CRAFT) to curate task-specific
samples from raw data for a wide variety of tasks. CRAFT
only requires a small set of few-shot examples from a
user to initiate the process of
crawling and structuring task examples. CRAFT
first detects relevant corpus examples from
large-scale unannotated corpora using similarity-based
retrieval. Then it uses LLMs to structure these examples
into a proper task format, effectively transforming
free-text documents into custom-formatted task samples
for fine-tuning.

We demonstrate the effectiveness of CRAFT on four diverse
tasks: three
QA tasks -- in biology, medicine and
commonsense --  as well as a text
summarization generative task. Our results show that models
fine-tuned on CRAFT-generated datasets achieve performance
that is either better than or comparable to
instruction-tuned LLMs.
Moreover, CRAFT not only outperforms other fully synthetic
data generation methods, such as
Self-Instruct~\citep{wang2023self} and
Evol-Instruct~\citep{xu2023wizardlm}, but also exhibits
robustness to variations in the quality of the initial few shots.
This holds across diverse tasks,
LLMs, and dataset sizes, highlighting the effectiveness of
our approach. We publicly release the code to craft datasets
for other tasks as well as all datasets and checkpoints at
\href{https://github.com/ziegler-ingo/CRAFT}{github.com/ziegler-ingo/CRAFT}.

\section{Related Work}
\subsection{Optimizing LLMs for Specific Tasks}
\paragraph{Prompting:}
Prompts are added to the input to provide additional context that guides the computation and output of a model~\citep{gu2023systematic}. A prompt usually takes the form of a natural language instruction~\citep{radford2019language,brown2020language}.
Prompting is commonly used with instruction-tuned models to define tasks and extract responses from language models, using natural language, without gradient updates.

\paragraph{Zero-Shot Inference:}
Originally discovered in the vision domain, zero-shot inference~\citep{larochelle2008zero} is a technique that allows models to generalize their learned knowledge from pre-training to previously unseen classes, tasks, or sample instance variants at inference time without gradient updates.
Pre-training LLMs on large corpora produces semantic representations that are generally applicable to multiple downstream tasks.
GPT-2~\citep{radford2019language} demonstrated that the
acquired capabilities can then be activated by prompting a new task in natural language.
However, zero-shot inference often falls short of the performance achieved by few-shot learning~\citep{brown2020language}.

\paragraph{Few-Shot Learning:}
In few-shot learning, the model is provided with a small
number of task-specific examples at inference time.
The few-shot examples are given to the model in the prompt, in a technique known as in-context learning~\citep{brown2020language}.
While full fine-tuning generally requires a substantial amount of labeled data, few-shot learning offers an inexpensive alternative to adapt a model to a new task with a limited number of examples~\citep{dong2024surveyincontextlearning}.
Nonetheless, few-shot learning faces several challenges,
including inaccurate assessment of the underlying data
distribution~\citep{song2023comprehensive}, biases related
to small sample sizes~\citep{song2023comprehensive}, and
sensitivity to shot length~\citep{liu2024lost}, shot quality and noise \citep{perez2021true,chang2021selectgen,chen2022noise}.

\paragraph{Full Fine-Tuning:}
During full fine-tuning, all model parameters are updated on
a large dataset with the goal of
adapting the model to a domain, task or dataset \citep{howard-ruder-2018-universal}.
This approach usually provides the best performance by learning task-specific patterns and relationships that may not be captured by pre-training and zero- or few-shot learning alone.
However, it requires a dataset of appropriate size.

\paragraph{Instruction Tuning:}
Instruction tuning~\citep{wei2022finetuned} is a type of full fine-tuning that optimizes a model to produce more relevant answers to questions or instructions~\citep{leike2018scalable,askell2021general}.
This approach enables language models to understand and follow user intents rather than simply continuing the input text.
Instruction-tuned models produce answers preferred by humans for tasks ranging from question-answering to summarization~\citep{ouyang_2022}.
The challenge is obtaining a large high-quality dataset that is both task-specific and in the desired instruction-output format.

\paragraph{Low-Rank Adaptation:}
Full fine-tuning may be too expensive for LLMs but the
difference between pre-trained weights and their fine-tuned
counterparts often has low rank~\citep{li2018measuring,aghajanyan2021intrinsic}.
Low-Rank Adaptation~\cite[LoRA]{hu2021lora} approximates these low-rank matrices during fine-tuning, and is efficient because it freezes the full model and only learns the low-rank matrices, which typically results in learning the equivalent of 2\% of the model's parameters.

\subsection{Synthetic Data Generation}
Synthetic data refers to artificially generated data that mimics the characteristics of real-world data~\citep{little1993statistical}.
It can be generated using statistical~\citep{sue1987linear,maqsud2015synthetic} or deep neural approaches~\citep{sutskever2011generating} with the aim of replicating the patterns, distributions, and structures found in real-world datasets.

\paragraph{Fully Synthetic Data Generation:}
A dataset is fully synthetic if the question or instruction, the possible context, as well as the answers are generated synthetically.
Methods such as Self-Instruct~\citep{wang2023self} and Evol-Instruct~\citep{xu2023wizardlm} generate general-purpose datasets by prompting LLMs to create task instructions and corresponding outputs.
Other approaches focus on generating task-specific fine-tuning data through the rephrasing of existing datasets~\citep{yin2023dynosaur,gandhi2024better} or on synthesizing pre-training data from general-purpose corpora~\citep{maini2024rephrasing}.
When applied to fine-tuning, these methods are either implemented via complex, resource-intensive multi-agent workflows~\citep{mitra2024agentinstruct} or are constrained to a narrow set of tasks because the generation process relies on models fine-tuned for those specific tasks~\citep{nayak2024learning}.

Two primary challenges of current approaches to fully synthetic data generation are repetition and inconsistent data quality.
Evaluations have indicated that many samples in fully synthetic datasets tend to exhibit high similarity to one another or to the seed samples~\citep{honovich-etal-2023-unnatural,wang2023self}.
Moreover, reported correctness rates for these datasets suggest that a substantial portion of the generated samples may not meet desired correctness standards~\citep{chen2024alpagasus}.
These observations highlight areas where fully synthetic data generation can be refined.

\begin{figure*}[htpb]
    \centering
	\includegraphics[scale=0.645]{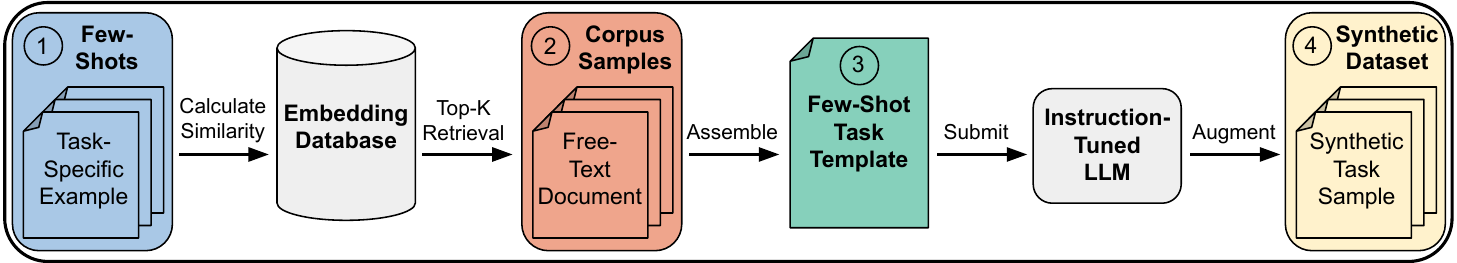}
        \caption{%
        Synthetic dataset generation process.
        Given a small set of task-specific few-shots \circled{1}, we retrieve the top-k most similar free-text documents \circled{2} from an embedding database.
        Each document is then integrated into a task template \circled{3} alongside original few-shots and an instruction prompt.
        An instruction-tuned LLM generates new synthetic task samples \circled{4} by augmenting the content of the corpus samples to mimic the style of the few-shots.
        The transformation process for each numbered step is illustrated with example documents in Figure \ref{fig:corpus-to-task_sample}.
        \label{fig:pipeline}
        }		
\end{figure*}

\paragraph{Partially Synthetic Data Generation:}
In partially synthetic data generation, a portion of the input, context, or output is generated synthetically, while the remaining portion is human-curated.
It is distinct from approaches that combine fully synthetic and purely human-curated samples at the dataset level, such as Phi~\citep{gunasekar2023textbooks}.

One recent approach is reverse instruction
generation~\citep{koksal2023longform}, where a language
model,
provided with a human-curated output in context,
generates
the instruction that would have prompted this output.
This produces more coherent and
correct input-output pairs because the LLM does not need to
generate the longer and more complex component of the data
sample. There are also approaches where, conversely, the
output is synthetically generated from human-curated input
samples.
Such methods employ distillation to extract patterns from larger models to teach those patterns and skills to smaller
models~\citep{mukherjee2023orca,mitra2023orca2}.

Partially synthetic data generation can alleviate some of the quality and diversity concerns inherent to fully synthetic approaches. However, using a raw corpus document as the output may introduce noisy or unnecessary information~\citep{agarwal2007much}. Data augmentation has been shown to mitigate these issues for pre-training data generation~\citep{maini2024rephrasing}. When applied to fine-tuning data, though, such augmentation typically requires a powerful model (e.g., GPT-4~\citep{openai2023gpt4}) to construct an intermediate synthetic dataset for fine-tuning a sample creator model~\citep{chen2024dog}. This multi-stage approach can lead to a sample creator model that essentially distills the larger model’s knowledge, potentially limiting task flexibility based on the synthesized training data. 
Alternatively, some methods have started to use retrieval to enhance diversity:
SynthesizRR~\citep{divekar-durrett-2024-synthesizrr} uses knowledge distillation from larger models while incorporating retrieved documents as in-context examples.
RADA~\citep{seo2024rada} relies on already structured input-output pairs from other existing datasets and retrieves them to improve the generation prompt for low-resource tasks.

In contrast, CRAFT streamlines this process by operating directly on unstructured, free-text corpora. Our approach produces fully synthetic data but leverages the quality and diversity advantages of human-written documents from partially synthetic data generation approaches while removing noise through augmentation.
CRAFT does not require intermediate datasets, nor a
separately fine-tuned model, nor knowledge distillation from
a larger model; instead, it relies only on a small number of human-curated examples, retrieval, and in-context learning.

\section{The CRAFT Approach}
\subsection{Architecture Overview}
CRAFT is used to fine-tune language models by generating task-specific synthetic datasets, given a few human-curated examples of the task.
During CRAFT (see Figure \ref{fig:pipeline}), we retrieve
human-written, free-text documents from a large collection
of corpora by calculating their similarity to the provided
few-shots and transforming them into
the task-specific format
through augmentation.
The only human effort required is in writing a small number of high-quality examples of the target task.
CRAFT has two phases: In the initial phase, an embedding database is created from large corpora. 
While this phase can be resource-intensive, its cost is incurred only once for all subsequent tasks, and it can be easily expanded with new corpora.
In the second phase, the user-generated, task-specific few-shot examples are embedded,
enabling the retrieval of relevant documents by calculating similarity measures between few-shots and corpus documents.
Once relevant documents are retrieved,
an instruction-tuned LLM is used to
augment the retrieved free-text documents into a
task-specific design, generating synthetic task samples in
the layout that is needed for instruction-tuning (illustrated in Figure \ref{fig:pipeline}).
Finally, the synthetic dataset is used to fine-tune a task-specific language model.
We report implementation details for the whole CRAFT framework in Appendix \ref{sec:implementation_details}.

\begin{figure*}[htpb]
\centering

\begin{tcolorbox}[
    colback=textbox,
    colframe=headerbox,
    colbacktitle=headerbox,
    title=\circled{1} \hspace{0.1cm} Few-Shot Design,
    fonttitle=\bfseries\color{black},
    fontupper=\fontsize{10.5}{12.6}\selectfont
]
   \textbf{Text:} \textcolor{cblue}{However, it has become clear that human chromosomes also carry a great deal of information} that is epigenetic, and not contained in the sequence of the DNA itself. Imprinting is one example. Another is seen in the phenomenon of mono-allelic expression, in which only one of the two copies of certain human genes is expressed.
   \vspace{0.15cm}
   \newline
   \textbf{Question:} What is epigenetic inheritance, and what are two examples of epigenetic changes? \vspace{0.15cm}\newline
   \textbf{Options:} A. Epigenetic inheritance signifies any heritable difference in the phenotype [\dots] \newline
   B. Epigenetic inheritance refers to inheriting variations in the number of chromosomes [\dots] \newline
   C. Epigenetic inheritance implies inheriting acquired traits during lifetime, whereas two [\dots] \newline
   D. Epigenetic inheritance denotes acquiring beneficial mutations via natural selection, and [\dots] \vspace{0.15cm}\newline
\textbf{Answer:} A. 
\end{tcolorbox}

\begin{tcolorbox}[
    colback=textbox,
    colframe=headerbox,
    colbacktitle=headerbox,
    title=\circled{2} \hspace{0.1cm} Corpus Sample,
    fonttitle=\bfseries\color{black},
    fontupper=\fontsize{10.5}{12.6}\selectfont
]
   \textbf{Text:} \textcolor{corange}{Proteins are involved in the formation of higher-ordered chromosome structures, such as} chromosome loops. Some proteins, including special AT-rich sequence-binding protein-1 (SATB1), CCCTC-binding factor (CTCF) and cohesin, play key roles in disease development and recovery.
\end{tcolorbox}

\begin{tcolorbox}[
    colback=textbox,
    colframe=headerbox,
    colbacktitle=headerbox,
    title=\circled{3} \hspace{0.1cm} Few-Shot Task Template,
    fonttitle=\bfseries\color{black},
    fontupper=\fontsize{10.5}{12.6}\selectfont
]
   \textcolor{cgreen}{\texttt{<s>[INST]}} \textcolor{cmint}{Please carefully read the text below. Then, generate exactly one question along with} four answer choices designated as A, B, C, and D based on the provided text. Then, respond to the question with the correct answer using only the corresponding letter label. Return the output only as a JSON structure in this format: \texttt{\{"question":} [\dots]\texttt{,} \texttt{"options":} [\dots]\texttt{,} \texttt{"answer":} [\dots]\texttt{\}} \newline
   \textcolor{cblue}{However, it has become clear that human chromosomes also carry a great deal of [\dots]} \textcolor{cgreen}{\texttt{[/INST]}} \texttt{\{"question":} [\dots]\texttt{,} \texttt{"options":} [\dots]\texttt{,} \texttt{"answer":} [\dots]\texttt{\}}\textcolor{cgreen}{\texttt{</s>}} \vspace{0.2cm} \par
   \hfill \textit{Repeat for randomly sampled few-shots 2 and 3} \hfill \vspace{0.2cm} \newline
   \textcolor{cgreen}{\texttt{<s>[INST]}} \textcolor{cmint}{Please carefully read the text below. Then, generate exactly one question along [\dots]} \newline
   \textcolor{corange}{Proteins are involved in the formation of higher-ordered chromosome structures, [\dots]} \textcolor{cgreen}{\texttt{[/INST]}}
\end{tcolorbox}

\begin{tcolorbox}[
    colback=textbox,
    colframe=headerbox,
    colbacktitle=headerbox,
    title=\circled{4} \hspace{0.1cm} Synthetic Task Sample,
    fonttitle=\bfseries\color{black},
    fontupper=\fontsize{10.5}{12.6}\selectfont
]
   \textbf{Question:} Which proteins play key roles in the formation of higher-ordered chromosome structures and disease development? \vspace{0.15cm}\newline
   \textbf{Options:} A. SATB1, CTCF, and cohesin \newline
    B. Histone proteins only \newline
    C. Transcription factors and co-TFs \newline
    D. RNA polymerase II and transcription factors \vspace{0.15cm}\newline
\textbf{Answer:} A.
\end{tcolorbox}
\vspace{-0.2cm}
\caption{%
Step-by-step synthetic task sample generation process for BioQA.
The color coding indicates where each section is reused throughout the process.
For readability, we shorten text sections in this figure,
indicated by ``[\ldots]''.
\circled{1} Few-shot design: the layout of a user-written few-shot sample that is used to guide the retrieval and task sample creation process.
\circled{2} Corpus sample: a retrieved free-text document from the embedding database based on cosine similarity to the user-written few-shot.
\circled{3} Few-shot task template: the prompting template that is used to augment the retrieved corpus sample into a synthetic task sample by using multiple few-shots as in-context examples.
\circled{4} Synthetic task sample: this is an actual synthetic task sample that is generated from the corpus sample \circled{2} using the few-shot task template \circled{3}.
}
\label{fig:corpus-to-task_sample}
\end{figure*}

\subsection{Few-Shot Examples}
A small number of human-curated few-shots serve as the
``definition'' of the task, i.e., they indicate how the task
is to be performed.
The few-shot samples consist of three elements:
(i) a long text that mirrors in language, content, and accuracy what a high-quality corpus sample from the web should look like,
(ii) a natural language instruction for the task to be performed, which can take the form of a direct instruction or a question about the text, and 
(iii) an output that satisfies the instruction or answers the question the way that the final model should later respond.
Length statistics for texts, instructions, and outputs of our few-shots can be found in the XS row of Appendix \ref{sec:dataset_stats}.

We note that the task does not need to be explicitly
specified. For example, there is no need to state the task
as ``biology question-answering''; it is sufficient for the
human-curated few shots to focus on QA in the domain of
biology.
If multiple-choice questions or single-letter outputs are in the few-shots, this will result in a corresponding dataset and fine-tuned model behavior.
These examples show that CRAFT is highly customizable: Few-shot
examples enable users to tailor the model’s behavior to
specific formats, use cases, or domains. Users can create
few-shots with unique terminology, style preferences, or
domain-specific constraints, optimizing the retrieval and
the final model’s performance for particular tasks.

\subsection{Corpora and Embedding Database}
The embedding database is a key element of CRAFT as it
provides, for all corpora, embeddings of human-written documents that should be retrievable for task-specific augmentation.
It is, therefore, important that the embedding database encompasses a wide variety of linguistically and semantically diverse documents.
This diversity can be achieved by including corpora that exhibit different writing styles, tones, and vocabularies.
Task-specific, task-agnostic, public, and also private documents can provide a comprehensive coverage of relevant information.
The more varied the documents in the embedding database, the
better the coverage
will be for diverse or rare tasks.
Notably, CRAFT can also handle sensitive company data, as the encoding, storage, and retrieval can be performed on-site.

\subsection{Document Retrieval}
\label{sec:corpus_retrieval}
Our retrieval system is task agnostic, both in terms of domain and complexity, in contrast to previous approaches~\citep{ein2020corpus,dai22a-dialog-inpainting,lewis-etal-2021-paq}.
The CRAFT approach relies on human-curated few-shot examples as query documents and can dynamically retrieve any document of the base corpora.
As the few-shot samples include a text containing the
domain, the instruction or question, as well as the output,
the resulting embedding representation of the sample
contains contextualized~\citep{reimers2019sentence} semantic
information about both the domain and the nature of the task to be performed.
Relevant text documents that contain similar latent features as the few-shots are retrieved from the corpora by calculating similarity scores based on the embedded few-shots and corpus samples.

As corpus size increases, the risk of retrieving redundant or similar corpus samples also increases.
This is partly due to the growing volume of documents, but also because the diversity of documents within the corpora may plateau, resulting in a higher proportion of similar documents. Designing few-shots that are sufficiently diverse in topic may alleviate this issue.
For example, when creating few-shots for biology question-answering, various subtopics of biology, such as genetics, anatomy, or physiology, should be covered to broaden the range of retrieved documents.

\subsection{Task Sample Synthesis}
The retrieved documents naturally contain noise~\citep{agarwal2007much} and lack the formatting required for fine-tuning.
Therefore, it is necessary to convert these free-text
documents into appropriate task samples
by removing noise and undesired sections. To address this, we utilize instruction-tuning prompt templates~\citep{sanh2022multitask,maini2024rephrasing} to augment free-text documents into task-specific training data while eliminating noise.
A few-shot task template consists of three elements:
(i) one or more few-shots,
(ii) a corpus sample, and
(iii) a brief instruction for the model to generate instruction-output pairs from the corpus sample.
The template 
structures all information from the instruction, the few-shot examples, and the retrieved corpus sample, which together
serves as input for the model that generates synthetic task samples. 

Figure \ref{fig:corpus-to-task_sample}, step 3, shows an example of how these templates guide the model in augmenting the corpus samples into synthetic task samples.
This augmentation step not only rephrases the text but also condenses the retrieved document down to the essential information required for the task.
This step produces final synthetic instruction-output pairs that can be used to fine-tune a language model.
Figure \ref{fig:corpus-to-task_sample}, step 4, shows an actual example output from the generated pool of synthetic training samples, and Appendix \ref{sec:dataset_stats} provides an overview of length statistics from the stages of corpus retrieval up to the synthesized input-output pairs.

\section{Experimental Setup}
This section summarizes how we implemented the CRAFT pipeline, the tasks on which we evaluate, dataset details for both CRAFT and the human-annotated baselines, model baselines and training details, as well as the evaluation metrics.

\subsection{CRAFT Implementation Overview}
We construct the embedding database from four large-scale corpora: C4~\citep{raffel2020exploring}, English Wikipedia, Stack Exchange~\citep{StackExchangeDataset}, and WikiHow~\citep{koupaee2018wikihow}.
After filtering documents by length (200-25,000 characters), this collection comprises 383 million documents.
We generate 384D embeddings for each document using SentenceTransformers~\citep{reimers2019sentence} with MiniLM~\citep{wang2020minilm} version \texttt{multi-qa-MiniLM-L6-cos-v1}.

During the retrieval phase, cosine similarity is computed between each few-shot example and the documents in each corpus shard (approx. 350K documents), retaining the top 5\% from each shard.
Then, a global top-k selection was performed over the retained candidates.
We employ a mixed strategy to balance topic coverage and specificity: 50\% of documents are retrieved based on their top-k cosine similarity to individual few-shot examples, while the other 50\% are retrieved based on similarity to the averaged embedding of all few-shots.

For task samples synthesis, we use Mistral 7B Instruct v0.2~\citep{jiang2023mistral} with temperature=0.7, top-k=40~\citep{fan2018hierarchical}, and top-p=0.9~\citep{holtzman2020curious}.
Input prompts used three randomly sampled few-shots.
To ensure data quality and structural integrity, we generate outputs as JSON objects, which allows for format validation.
Finally, we enhance dataset diversity by performing a deduplication step, filtering out generated samples that are highly similar to the seed few-shots or to each other using fuzzy string matching~\citep{ranjith2022review} with a token set similarity ratio of 0.85.
Further implementation details and filtering statistics are provided in Appendices~\ref{sec:implementation_details} and~\ref{sec:filtering_stats}, respectively.

\subsection{Tasks and Datasets}
To test the performance of the CRAFT pipeline, we evaluate it on five different tasks: multiple-choice (MC) biology QA, MC commonsense QA, MC medicine QA, summarization, and recipe generation. We first describe the details of CRAFT and then introduce the datasets for the human-annotated baselines.

\subsubsection{CRAFT Datasets}
We generate datasets for all tasks with sizes of 100, 500, 5,000, and 25,000. We refer to the few-shot datasets (with 8 or 32 examples) as size XS, and to the datasets with 100, 500, 5,000, and 25,000 examples as sizes S, M, L, and XL, respectively. For few-shot curation, we do not refer to any existing datasets; instead, the authors manually curate examples from corpora (see Appendix~\ref{sec:appendix-few-shot-design} for sources used).

\textbf{Multiple-Choice QA:}
 We generate three synthetic QA datasets in the domains of
biology (BioQA), commonsense (CSQA), and medicine
(MedQA). All datasets follow the MMLU multiple-choice
format~\citep{hendrycksmeasuring}, where each question is
accompanied by several answer options. Exactly one of these
options is correct. The task is to output the correct
answer, the letter label corresponding to the correct option.

\textbf{Generation:} We develop two synthetic datasets for the generation tasks of recipe generation (RecipeGen) and summarization. The goal of summarization is to convey accurate and concise information, while recipe generation focuses on creating coherent and structured text that adheres to specific formatting and stylistic conventions~\citep{wang2022learning}.

To build a synthetic summarization dataset, we first select a corpus sample and instruct the model to extract an extended section of text. In the second step, the extracted section is transformed into a summary, \textit{optionally incorporating elements from the raw text} (e.g., `abstract:', `conclusion:', or `TLDR'). This approach avoids using full corpus samples as the text to be summarized, as they can be lengthy and overly broad, potentially resulting in uninformative summaries. For recipe generation, our goal is to generate a list of ingredients and cooking steps for a specific recipe.

\subsubsection{Human-Annotated Datasets}
We select human-annotated datasets as optimal baselines for each task to compare the performance of models trained on CRAFT. For evaluation, we use the test split of these datasets, as detailed in \S\ref{sec:sub_eval_details}; therefore, the human-annotated datasets can also be considered in-domain.

\textbf{BioQA:}
We use the biology subset of ScienceQA~\citep{lu2022learn} with 1,192 training samples without images. ScienceQA sourced expert-curated question-answer pairs from online learning platforms, ensuring high quality and accuracy. The dataset's answer options range from two to five, include a single correct answer per question, and are randomized to prevent pattern recognition.

\textbf{MedQA:}
We use MedMCQA~\citep{pmlr-v174-pal22a} and randomly select 25,000 samples from the training split. The dataset comprises entrance exam questions from two of India's postgraduate institutions. All samples come from preparation materials or real exams created by medical professionals, with each question containing four answer options and one correct answer.

\textbf{CSQA:}
We use CommonsenseQA 2.0~\citep{talmor1commonsenseqa} and select 9,264 samples from the training split. The dataset was generated through a gamified yet controlled question generation process where players earned points by designing challenging yes/no questions that outperform an AI model. Generated questions were validated by other players, independent validators, and another model to ensure  they were well-formed, answerable, and representative of common sense.

\textbf{RecipeGen:}
We use RecipeNLG~\citep{bien-etal-2020-recipenlg} and select 25,000 samples from the training split. The recipes were scraped from cooking websites and post-processed through fine-grained cleaning and formatting to ensure correctness. Each recipe includes a title, ingredient list, and cooking steps. We exclude samples present in C4 based on provided URLs.

\textbf{Summarization:}
We use CNN-DailyMail~\citep{see-etal-2017-get} and randomly select 25,000 samples from the training split. This dataset is commonly used for summarization tasks as its CNN and DailyMail articles have highlights presented in abstract or bullet point formats.

\subsection{Baselines}
We compare CRAFT models trained on synthetic data against multiple baselines:

\noindent\textbf{Few-shot} is fine-tuned only on XS-size CRAFT datasets containing human-curated few-shot examples. This serves as our primary baseline, representing all human-curated data in our pipeline.

\textbf{Instruct:} Mistral 7B Instruct v0.2~\citep{jiang2023mistral} is instruction-tuned on proprietary instruction-following datasets. This provides a meaningful comparison to a similarly-sized instruction-tuned model, though trained on undisclosed data of unknown quality and quantity. Surpassing this baseline would indicate CRAFT's ability to generate high-quality synthetic data.

\textbf{In-Domain (ID)} is fine-tuned on human-curated training splits from evaluation datasets. This baseline represents the optimal performance achievable with human quality datasets.

We additionally compare CRAFT against two popular synthetic data generation methods:

\textbf{Self-Instruct}~\citep{wang2023self} distills 
new task samples from instruction-tuned models, typically using fewer than 10 seed examples per task.

\textbf{Evol-Instruct}~\citep{xu2023wizardlm} generates new samples by rewriting seed examples to be more complex, to incorporate more domain-specific 
concepts, or to add step-by-step reasoning.

We initialize these methods with 8 few-shot examples to follow Self-Instruct's setup and also generate four dataset sizes (S: 100, M: 500, L: 5,000, XL: 25,000) to follow CRAFT's approach. 
We use each method's original hyperparameters and the same seed examples.

\subsection{Training and Optimization}
\label{sec:implementation_details_training}
We fine-tune base models using either CRAFT datasets or
human-annotated datasets. Models fine-tuned with
human-annotated datasets are denoted with the suffix -ID
(in-domain), while models fine-tuned with CRAFT datasets are
labeled as -CRAFT\textsubscript{\#}, where \# represents the
training split size (XS, S, M, L, or XL). Our primary
experimental setting uses Mistral 7B v0.2 as the base model
with 32 few-shot examples in CRAFT. Additional experiments
may use 8-shot settings or another base model, Llama 3 8B,
specified explicitly when used. We also evaluate (as
baselines) instruct versions of base models without fine-tuning.

For all experiments, low-rank adaptation~\citep[LoRA]{hu2021lora} fine-tuning is performed using 16-bit BrainFloat~\citep{abadi2016tensorflow} as the computation type.
All implementations use PyTorch~\citep{paszke2019pytorch} and HuggingFace libraries~\citep{wolf2020transformers}.
For optimization, the adaptive momentum optimizer with decoupled weight decay~\citep{loshchilov2019adamw} of 5\% and a learning rate of $1 \times 10^{-4}$ is employed.
A linear learning rate schedule is applied with a warm-up ratio of 10\%.
Models are fine-tuned for three epochs across all tasks and dataset sizes for CRAFT. When training only on human-curated few-shots, we adopt a batch size of 2; otherwise, we use batch size 16 or equivalent gradient accumulation steps.
Following~\citet{dettmers2024qlora}, we apply LoRA adapters to every linear layer (query\hbox{-}, key\hbox{-}, value\hbox{-}, output\hbox{-}, gate\hbox{-}, up\hbox{-} and down-projection matrices) with rank 64 and $\alpha$ 64 and 0.1 dropout rate. Bias terms in update matrices are deactivated.
This configuration adds 2.3\% (160 million parameters) to the base model's 7 billion parameters as LoRA adapters. Frozen base parameters remain unchanged during training, with updated parameters merged post-training.

\begin{figure*}[t]
    \centering
    \includegraphics[width=0.98\linewidth]{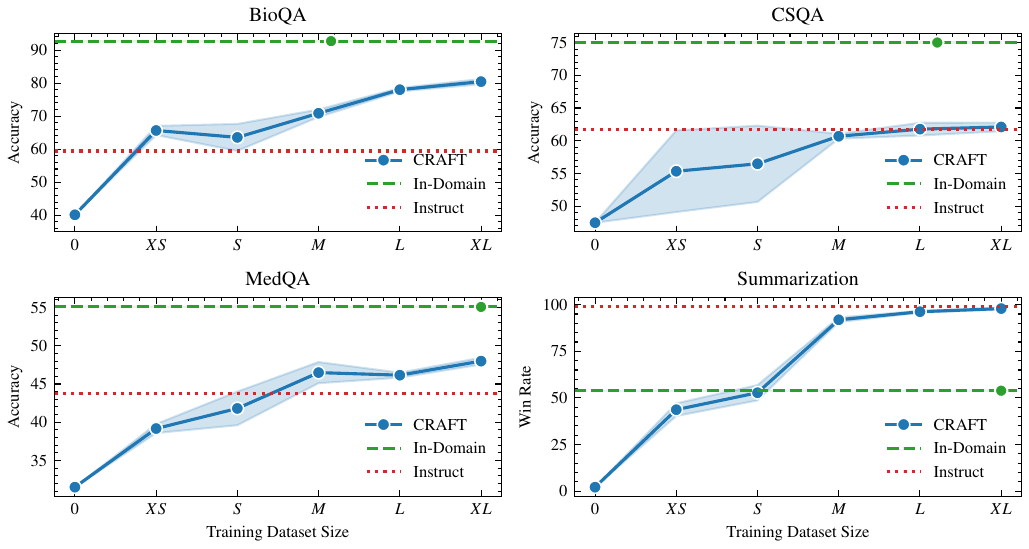}
    \caption{Performance scaling with increasing data size across multiple tasks using CRAFT with 32 few-shot examples.
Graphs demonstrate consistent improvements as training data grows from few-shot (XS) to 25,000 synthetic samples (XL). CRAFT models consistently match or exceed Instruct performance (dotted red line). Shaded regions indicate standard deviation across three runs.}
    \label{fig:main_results}
\end{figure*}

\subsection{Evaluation}
\label{sec:sub_eval_details}
\subsubsection{Metrics}

\textbf{QA Tasks:} 
We evaluate multiple-choice QA tasks using \textit{accuracy}, following MMLU's approach of assessing logarithmic probabilities for vocabulary tokens corresponding to answer labels~\citep{hendrycksmeasuring}. We perform greedy decoding without temperature scaling across answer choices ranging from A-B to A-E.

\textbf{Generation Tasks:}
While automated metrics like ROUGE~\citep{lin-2004-rouge}
and METEOR~\citep{banerjee-lavie-2005-meteor} are efficient, their reliance on n-gram overlap limits effectiveness for generative tasks~\citep{barbella2021comparison}. Reference text quality issues and length disparities further reduce reliability~\citep{graham2015re,sai2019re,celikyilmaz2020evaluation}. For example, \citet{sottana2023evaluation} found human reviewers often rank benchmark answers among the worst options.

We instead evaluate generations using \textit{LLMs as judges}~\citep{eldan2023tinystories}, where models provide binary preference scores for output pairs, yielding win rates as metrics~\citep{chiang2024chatbot}. This approach shows high inter-rater reliability comparable to human annotators~\citep{hackl2023gpt4rater,sottana2023evaluation,liu2023g}.

For general-purpose outputs, we use the popular Alpaca-Eval
benchmark~\citep{alpaca_eval}
that evaluates
multiple LLMs on about 650 human-curated questions~\citep{dubois2023alpacafarm}. We select Llama 3 70B~\citep{dubey2024llama3herdmodels} as our annotator model due to its open nature and cost-efficiency for high-volume experiments. As of January 2025, Llama 3 70B ranks 4th in human agreement with a score of 67.5, close to customized GPT-4 versions at 69.2.

\subsubsection{Datasets}

We benchmark all models on the original test splits of the human-annotated datasets used for the in-domain (ID) baselines:

\textbf{BioQA:} 397 test samples without images from ScienceQA's biology subset~\citep{lu2022learn}.

\textbf{MedQA:} 4,183 validation samples from MedMCQA~\citep{pmlr-v174-pal22a}.

\textbf{CSQA:} 2,541 validation samples from CommonsenseQA 2.0~\citep{talmor1commonsenseqa}.

\textbf{RecipeGen:} 1,000 high-quality samples from RecipeNLG~\citep{bien-etal-2020-recipenlg}.

\textbf{Summarization:} 1,000 test samples from CNN-DailyMail~\citep{see-etal-2017-get}.

\section{Results}
\label{sec:results}

\subsection{Scaling the Data}
Figure \ref{fig:main_results} shows performance improvements from data scaling, 
reporting means and standard deviations over three random seeds. 
We observe consistent gains across four tasks with increasing data size relative to
the few-shot baseline: 22\% (from 65.7 to 80.4), 12\% (from 55.3 to 62.1), 23\%
(from 39.1 to 48.0), and 124\% (from 43.7 to 97.9) for
BioQA, CSQA, MedQA, and Summarization, respectively, from XS to XL.
These results demonstrate CRAFT's effectiveness across
diverse tasks starting from minimal curated examples. Models show appropriate scaling from 100 to 25,000 synthetic samples. Additionally, 
models trained with fewer examples (32, 100) exhibit
higher variance than those trained with 5,000 and 25,000
examples, as indicated by the shaded regions that
visualize the standard deviation in the plots.

Notably, CRAFT matches or exceeds Instruct performance across all tasks except RecipeGen.\footnote{We further 
investigate the RecipeGen results
in \S\ref{sec:recipegen}.} It is worth
noting that CRAFT uses an LLM in a limited way (to
restructure and rewrite existing corpora) that seems to
exclude the possibility that distillation has played a
role. However, even if distillation were to be considered the
reason for good CRAFT performance, the results indicate
otherwise:
we use the same model as Instruct, Mistral 7B Instruct v0.2,
to paraphrase existing corpora examples but
achieve even stronger results.

Finally, we observe that
CRAFT models outperform the In-Domain (ID) baseline of 25,000 samples in summarization. 
For other
tasks, while we observe lower performance than ID with sample numbers between 1,192 for BioQA, 9,264 for CSQA, and 25,000 for MedQA, 
we speculate that this could be due to in-domain
evaluation for human-curated data. We use their
test split to evaluate our models, which may give these
models an unfair advantage. We investigate this further in
the next section.

\subsection{Data Contamination and OOD Generalization}
\label{sec:results-ood-generalization}

For the experiments
in Figure \ref{fig:main_results},
we investigate potential data contamination between test and training examples. We conduct 5-gram weighted
Jaccard similarity analyses between CRAFT or in-domain (ID) datasets and the test dataset.
For each sample, we combine the instruction and output and gather 5-gram
frequencies for the whole dataset.
We then calculate the Jaccard similarity between the 5-gram
frequency distributions of the respective CRAFT/ID and test
datasets, where n-grams receive weight proportional to their frequency.

\begin{table}[t]
    \centering
    \resizebox{0.5\textwidth}{!}{
    \begin{tabular}{lcc}
        \toprule
        Dataset & In-Domain (ID) & CRAFT\textsubscript{XL} \\
        \midrule
        ScienceQA (In-Domain) & \textbf{92.7}  & 80.4  \\
        \midrule
        MMLU\textsubscript{Medical Genetics}     & \textbf{67.0} & 66.0 \\
        MMLU\textsubscript{Anatomy}              & 58.5 & \textbf{60.0} \\
        MMLU\textsubscript{High School Biology}  & 68.1 & \textbf{69.0} \\
         MMLU\textsubscript{College Biology}      & 69.4 & \textbf{70.8} \\
        \midrule
        MMLU-Avg & 65.8 & \textbf{66.5} \\
        \bottomrule
    \end{tabular}
    }
    \caption{Out-of-domain performance of in-domain (ID) vs CRAFT\textsubscript{XL} models on biology QA tasks. While ID outperforms on its in-domain ScienceQA test set, CRAFT\textsubscript{XL} shows better generalization to three of four OOD biology subsets from MMLU.}
    \label{tab:cross_domain}
\end{table}

\begin{figure}[t!]
    \centering
    \includegraphics[width=0.95\linewidth]{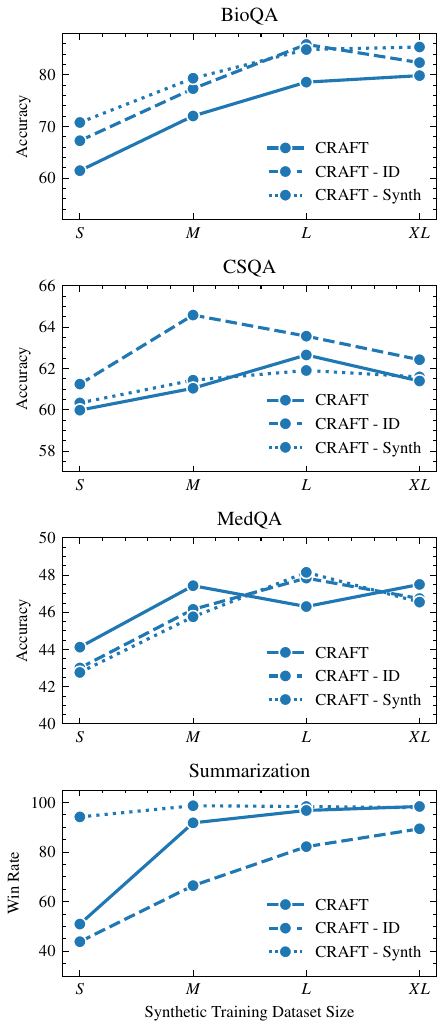}
    \caption{Performance comparison when CRAFT's retrieval process is initiated with standard human-curated (CRAFT), in-domain (-ID) and purely synthetic (-Synth) few-shots. As the dataset size increases, performance converges across the different few-shot sources, indicating that the retrieval and augmentation framework of CRAFT effectively abstracts away the variability in the quality of the initial few-shots.}
    \label{fig:sensitivity_analysis-source}
\end{figure}

This analysis reveals that all CRAFT datasets have less than 0.4\%
similarity with the test sets, whereas the in-domain datasets 
show much higher similarities: BioQA (16.6\%, 1,192 samples), CSQA (4.4\%, 9,264 samples),
MedQA (1.1\%, 25,000 samples), and Summarization (0.3\%, 25,000 samples), indicating some overlap
between train and test splits.
However, the substantial overlap in
in-domain datasets suggests that their reported performance 
might benefit from train-test similarity.

To isolate generalization capabilities, we evaluate the 
in-domain (ID) baseline and CRAFT models on four 
out-of-domain (OOD) biology QA tasks from MMLU~\citep{hendrycksmeasuring}. 
Table~\ref{tab:cross_domain} shows that while the ID model outperforms 
CRAFT\textsubscript{XL} by 12.3 percentage points on its in-domain test set,
CRAFT\textsubscript{XL} performs better on three of the four OOD biology tasks. 
This indicates that CRAFT's training datasets offer better generalization 
capability and robustness across different domains than human-annotated datasets.

\subsection{Sensitivity Analyses}
We investigate the sensitivity of the CRAFT pipeline based on different setups: the source of few-shots, 
the number of few-shots, and the base model
used for fine-tuning.

\begin{figure}[t]
    \centering
    \includegraphics[width=0.93\linewidth]{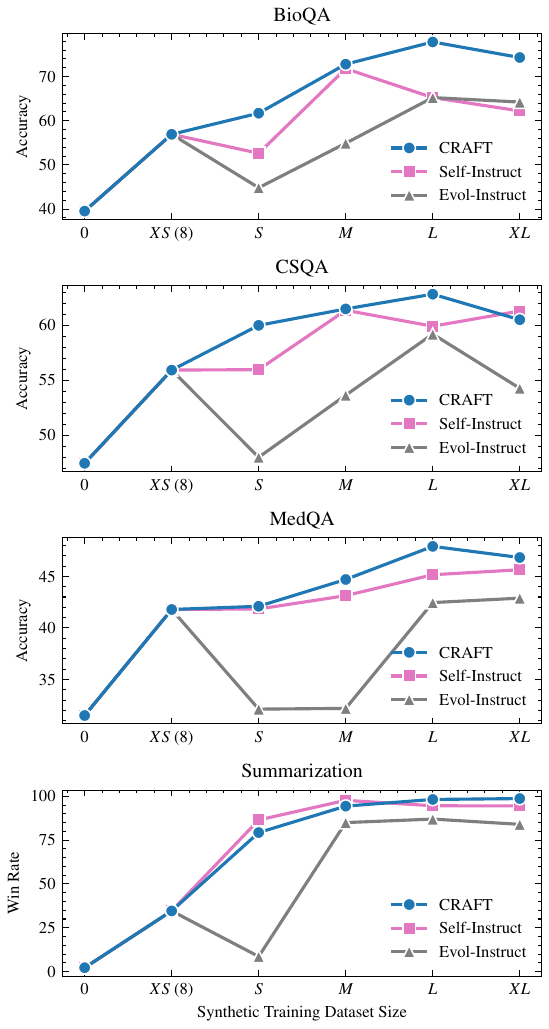}
    \caption{Performance of CRAFT versus Evol-Instruct and Self-Instruct across tasks and dataset sizes with 8 few-shots. CRAFT shows better scaling and higher accuracy than both baselines in most settings.}
    \label{fig:synthetic_baselines}
\end{figure}

\subsubsection{Few-Shot Source}

To assess the sensitivity of CRAFT to the source of the initial few-shots, we evaluate two additional variants alongside our standard human-curated few-shots. In the in-domain (\textbf{ID}) setting, we directly select few-shots  from the training splits of our evaluation datasets and manually pair them with appropriate web texts following the design illustrated in Figure~\ref{fig:corpus-to-task_sample}. This setup represents an optimal distribution match to the test set. The synthetic (\textbf{Synth}) setting employs zero-shot prompts using Mistral 7B Instruct v0.2 to generate both the few-shots and the corresponding web texts without further curation. Although this approach minimizes human effort, it carries the risk of producing repetitive or lower-quality examples.

We run the full CRAFT pipeline using these few-shot variants, fine-tuning Mistral 7B v0.2 and evaluating performance on our standard benchmarks. As shown in Figure~\ref{fig:sensitivity_analysis-source}, most tasks yield very similar performance regardless of the few-shot source. For CSQA and MedQA, performance varies by at most 1.02 and 0.96 percentage points (pp.), respectively, confirming the stability of our approach. Summarization exhibits higher variance across few-shot sources, with a 9 pp. difference favoring manually curated CRAFT few-shots. However, when scaled to CRAFT\textsubscript{XL}, different few-shot sources lead to more similar results, despite the initially higher variability. Overall, these findings highlight the robustness of CRAFT, demonstrating that even with synthetic few-shots, it consistently generates high-quality, task-specific datasets.

\begin{table}[t!]
\centering
\resizebox{\columnwidth}{!}{%
  \begin{tabular}{lcccc}
    \toprule
    Num FS & BioQA  & CSQA   & MedQA  & Summ \\
    \midrule
    8  & 74,31 & 60,49  & 46,83 & 98,70 \\
    32 & 79,85 & 61,39  & 47,50 & 98,40 \\
    \bottomrule
  \end{tabular}%
}
\caption{Accuracy and win rate of CRAFT\textsubscript{XL} when initiated with 8 and 32 human-curated few-shots. Using just 8 few-shots to initialize CRAFT achieves comparable performance in three of four tasks.}
\label{tab:8-32-comparison}
\end{table}

\subsubsection{Number of Few-Shots}

To better understand the impact of initial human effort required on our synthetic data generation process, we investigate the sensitivity of CRAFT to the number of few-shot examples provided.

Table~\ref{tab:8-32-comparison} compares the performance of CRAFT\textsubscript{XL}, which generates 25,000 synthetic samples, when our retrieval started with 8 versus 32 human-curated few-shots.
The results indicate that, while increasing the number of few-shots can yield notable improvements (e.g., BioQA increases from 74.31 to 79.85), the overall performance across three of four tested tasks remains largely comparable.
In commonsense QA, medicine QA, and summarization, differences are minimal, with performances varying below 1 percentage point or even matching performance between the two settings.
These findings suggest that initializing CRAFT with as few as 8 few-shots is a valid and cost-effective option, significantly reducing the human effort required while still producing high-quality synthetic datasets.
Detailed sensitivity results for all dataset sizes (from XS to XL) are provided in Appendix~\ref{sec:additional-results-varying-few-shots}.

\subsubsection{Base Model}

In previous sections, we fine-tuned CRAFT models using the pretrained Mistral 7B model. Now, we repeat the experiments using the pretrained Llama 3 8B~\citep{dubey2024llama3herdmodels} model. We observe similar trends across all tasks, and the relative improvement is comparable when scaling up from few-shots to 25,000 examples, as illustrated in Table \ref{tab:model-ablation}.
Notably, the datasets created with Mistral 7B Instruct v0.2 are also effective in improving a model with a much stronger baseline performance, such as Llama 3 8B. This finding underscores that our framework leverages the sample-generating LLM only in a reduced capacity -- only to augment the retrieved documents -- without relying heavily on its inherent performance.

\begin{table}[t!]
\centering
\resizebox{\linewidth}{!}{
\begin{tabular}{llllll}
\toprule
Task & CRAFT\textsubscript{XS} & CRAFT\textsubscript{S} & CRAFT\textsubscript{M} & CRAFT\textsubscript{L} & CRAFT\textsubscript{XL} \\
\midrule
\textbf{BioQA} & 61.5 & 64.7& 68.8 & 73.0 & 78.3 \\
\textbf{CSQA} & 55.8 & 54.8 & 58.6 & 60.4 & 61.4 \\
\textbf{MedQA} & 49.5 & 49.6 & 52.2 & 51.5 & 53.2 \\
\textbf{Summ} & 37.3 & 32.7 & 86.6 & 96.8 & 96.9 \\
\bottomrule
\end{tabular}
}
\caption{Accuracy and win rate of CRAFT
  when Llama 3 8B is used as base model. Datasets generated using Mistral 7B v0.2 can further improve a stronger baseline model such as Llama 3 8B.}
\label{tab:model-ablation}
\end{table}

\subsection{Synthetic Data Generation Comparisons}
Figure \ref{fig:synthetic_baselines} shows the results when CRAFT is compared against other synthetic data generation methods.
CRAFT consistently outperforms Evol-Instruct (EI) across all tasks and 
dataset sizes. While Self-Instruct (SI) sometimes matches CRAFT's performance 
at smaller scales (S/M), CRAFT always achieves higher performance than 
SI at larger scales (L/XL), except for CSQA at the XL size. Moreover, 
CRAFT scales more consistently while performing better than the other methods. We believe that a fully distillation-based technique like SI could be competitive with CRAFT in some cases, especially if the number of few-shot examples were increased; 
however, the available corpus samples in CRAFT improve quality and diversity even with a small number of few-shots. 

Importantly, we observed significant differences in method sensitivity across few-shot sources. When running the methods with three different few-shot sources (human-curated, in-domain, and purely synthetic examples), CRAFT demonstrated notably lower variability. At the XL model size, CRAFT achieved an average standard deviation of 1.57 across all five tasks, compared to 3.53 for EI and 4.31 for SI, while also achieving better performance.

\subsection{Negative Results: Recipe Generation}
\label{sec:recipegen}
\begin{figure}[t]
    \centering
    \includegraphics[width=0.9\linewidth]{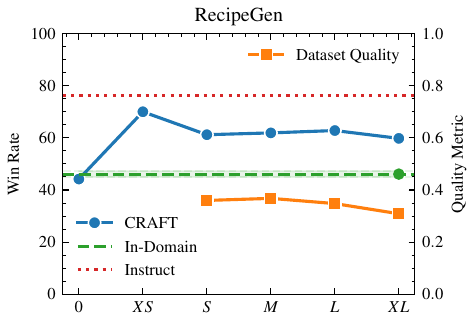}
    \caption{Non-scaling behavior in Recipe Generation: Dataset quality declines with increasing size (100 to 25K examples), showing an inverse scaling trend. This anomaly reflects diminishing data quality at scale.}
    \label{fig:negative_recipe}
\end{figure}

Out of our five tasks, we observe non-scaling behavior in one: Recipe Generation. While our manually curated few-shots are of high quality, we see a drop when scaling from 32 to 25,000 examples, as illustrated in Figure \ref{fig:negative_recipe}. CRAFT's performance is still better than the baseline with official human data, which means that the final dataset is usable. However, we explore why this reverse scaling occurs and examine the drop in performance.

\begin{table*}[t!]
\centering
\small
\resizebox{\textwidth}{!}{%
\begin{tabular}{@{}ll@{}}
\toprule
\textbf{Generation Method} & \textbf{Generated Sample} \\
\midrule

Human (ID) & \textbf{Sample 1:} Select the temperate deciduous forest ecosystem. \\
& A. This ecosystem has: long, cold winters and short, cool summers; many evergreen trees; soil that is poor in nutrients \\
& B. This ecosystem has: year-round rain and warm temperatures; soil that is poor in nutrients; many different types of organisms \\
& C. This ecosystem has: warm, wet summers and cold, wet winters; soil that is rich in nutrients; only a few types of trees \quad \textbf{Answer:} C \\
& \textbf{Sample 2:} Which organ controls the function of other body organs? \\
& A. skeleton \quad B. lungs \quad C. heart \quad D. brain \quad \textbf{Answer:} D \\
\midrule

CRAFT & \textbf{Sample 1:} Which organ plays the primary role in excreting acid and regulating electrolyte balance in the body? \\
& A. Lungs \quad B. Skin \quad C. Kidneys \quad D. Liver \quad \textbf{Answer:} C \\
& \textbf{Sample 2:} Which simple, single-celled organism formed the first stromatolites on Earth approximately 2 billion years ago? \\
& A. Cyanobacteria \quad B. Algae \quad C. Fungi \quad D. Protozoa \quad \textbf{Answer:} A \\
\midrule

Self-Instruct (SI) & \textbf{Sample 1:} Which of the following is a characteristic feature of prokaryotic cells? \\
& A. Nucleus and mitochondria are present in the cell. \quad B. The cell membrane is not present. \\
& C. The DNA is located in the nucleolus. \quad D. The cell wall is absent. \quad \textbf{Answer:} D \\
& \textbf{Sample 2:} Which of the following is a function of the endoplasmic reticulum (ER)? \\
& A. It acts as a site for protein synthesis and folding. \quad B. It functions in the storage and transport of lipids. \\
& C. It plays a role in the breakdown of complex carbohydrates. \quad D. It is involved in the process of cell division. \quad \textbf{Answer:} A \\
\midrule

Evol-Instruct (EI) & \textbf{Sample 1:} In the early stages of cellular evolution, how did organisms with rudimentary structures develop a complex and dynamic [\dots] \\
& In the pursuit of resource optimization, how did these ancient organisms establish an intricate signaling network, enabling [\dots] \\
& \textbf{Sample 2:} As you delve deeper into the intricate diaphyseal region of a human right femur, you'll encounter various fascinating [\dots] \\
& Further studies on the presence, distribution, and function of bone marrow [\dots] \\
\bottomrule
\end{tabular}%
}
\caption{%
Qualitative comparison of BioQA samples.
For each source, we show two examples to illustrate strengths or shortcomings.
CRAFT's outputs are well-formed and comparable to the human baseline.
Self-Instruct maintains the format, but Sample 1 contains a factual error, and Sample 2 includes multiple correct options, making Option B ambiguous in a single-answer context.
Evol-Instruct deviates from the task, failing to produce usable QA samples.}

\label{tab:qualitative-analysis}
\end{table*}

An initial analysis suggested that the CRAFT
pipeline tends to find less relevant examples over time. We conducted automated data
quality analysis to
analyze this on a larger scale. For 500 randomly sampled instructions from
different sizes of CRAFT datasets (i.e., the training sets), we used
Llama 3 8B Instruct  to answer the instructions. Then,
using Llama 3 70B Instruct as a judge, we compared win
rates, i.e., which output the model preferred: the gold
output in the CRAFT datasets or the output generated by
Llama 3 8B Instruct. We report the average win rate against the Llama outputs as the data quality metric. Higher scores
indicate that the pipeline created higher quality output
than Llama 3 8B Instruct's answers.

We observe a decline in data quality when scaling
to 25,000 examples: The 100- and 500-example sets 
achieve win rates of around 0.4, while the 25K set drops
to 0.3. This degradation likely causes the performance
decline during scaling. We also observe a similar trend in other
synthetic data generation methods like Self-Instruct and Evol-Instruct. 
While the final CRAFT dataset remains practical 
(outperforming the baseline using official human data), 
future work should incorporate stopping 
criteria or more quality validation.

\section{Qualitative Analysis}
To better understand the quantitative results presented in Section~\ref{sec:results}, we perform qualitative case studies by inspecting synthesized samples from multiple tasks and data generation methods.
This analysis provides intuition as to \textit{why} certain methods perform better than others and illustrates the style and patterns in generated samples.

\subsection{Case Study Design}
Our analysis includes both QA and generative tasks across samples generated by CRAFT, Self-Instruct (SI), Evol-Instruct (EI), and human annotators.
We use BioQA as the representative for QA tasks, but analyze samples from summarization and recipe generation separately due to format differences. 
After manually inspecting a wide range of outputs, we selected representative examples that either adhere to or violate the following three criteria:
(i) alignment with the expected task format,
(ii) content quality and factuality, as well as
(iii) stylistic appropriateness.
This setup illustrates common advantages and shortcomings.

\subsection{Question-Answering Observations}
\paragraph{Adherence to format:}
As Table~\ref{tab:qualitative-analysis} shows, both CRAFT and SI produce well-formed QA pairs in a concise format, matching the human in-domain data. 
EI often drifts into free-form essays, bundles multiple questions into one instruction, and often reaches maximum generation length before completing a sample.
The method's goal to evolve samples by increasing complexity leads to a breakdown in task adherence, which observably resulted in decreased performance in Section~\ref{sec:results}.

\paragraph{Domain fidelity and correctness:}
CRAFT leverages retrieved corpus passages to embed precise terminology (``\textit{stromatolites}'', ``\textit{electrolyte balance}’’) and produces correct QA pairs.  
Distractor options are semantically proximate yet unambiguously wrong, mirroring expert-curated style.  
SI's Sample 1 is factually incorrect, stating that the cell wall is absent in prokaryotes. This exemplifies the risk of ungrounded generation: the model produces plausible-sounding samples at the cost of factual integrity.
Furthermore, in SI's Sample 2, the distractor option B is also correct, creating annotation noise.
EI's questions are verbose and often include multiple sub-questions, making it unclear which question should or will be answered.

\subsection{Generative Tasks Observations}
\paragraph{Recipe Generation:}
The human-annotated in-domain recipes are terse, often just listing steps without providing descriptive details.

\vspace{0.5em}
\noindent\small\leftskip=0.5em\rightskip=0.5em
\emph{%
Add all dry ingredients together and mix well. Add remaining ingredients stirring only until moistened. [\dots]
}
\hfill{\footnotesize(Human (ID), Banana Nut Muffin)} \\
\par\leftskip=0pt\rightskip=0pt\normalsize

\noindent In contrast, CRAFT can produce specific, multi-step recipes with headings and detailed directions.

\vspace{0.5em}
\noindent\small\leftskip=0.5em\rightskip=0.5em
\emph{%
\textbf{Cook the apples:} In a saucepan, cook the sliced apples with sugar over medium heat until softened and caramelized. [\dots]  \\
\textbf{Add apple filling:}
Place the apple filling in a line down the middle of the pastry, leaving a 2-inch border on all sides. \\
\textbf{Brush the edges:} Brush the edges of the pastry with the beaten egg and water mixture.
} \\  
\textit{[\dots]} \hfill{\footnotesize(CRAFT, Apple Strudel)} \\
\par\leftskip=0pt\rightskip=0pt\normalsize

\noindent CRAFT's ability to generate more descriptive content stems from grounding the sample synthesis in more elaborate retrieval documents such as cooking blogs.
However, as CRAFT retrieves less relevant documents at larger scales, its outputs can degrade.
Instructions can become vague (e.g., \textit{``Prepare a dish using items from your well-stocked pantry, freezer, and fridge.''}) or convoluted, possibly from combining multiple separate recipes from a retrieved document into one (e.g., \textit{``Create a salad or bowl by combining greens, proteins, vegetables, cheeses, fruits, and nuts with the specified dressing.''}, resulting in an ingredient list that starts with greens but ends with \textit{``assorted fruits''} and a \textit{``scoop of ice cream''}. Complete recipe samples are shown in Appendix~\ref{sec:appendix-recipe_samples}.

\paragraph{Summarization:}
The human-annotated summaries are written in a headline-driven style, presenting information as extractive bullet points (e.g., \textit{``NEW: California Public Utilities Commission passed ban Thursday.''}.
While factually grounded, this format is limited in stylistic variety and can lack context as well as narrative coherence.
CRAFT learns to generate more abstractive, prose-style summaries that are tailored to specific instructional details 
by being exposed to varying retrieval contexts.
The full sample is provided in Appendix~\ref{sec:appendix-summarization_samples}.

\vspace{0.5em}
\noindent\small\leftskip=0.5em\rightskip=0.5em
\emph{%
\textbf{Instruction:} Summarize the text below, focusing on Bill Ackman's investment in Procter \& Gamble and his intentions for the company.
\textbf{Text:} [153 word text ...].
\textbf{Summary:} Bill Ackman, a hedge fund manager, believes that Procter \& Gamble's (P\&G) stock could be worth more than its current price due to poor marketing and pricing strategies. He holds around 1\% of the company's shares and is actively pushing for changes. [48 more words \dots]
}
\par\leftskip=0pt\rightskip=0pt\normalsize
\vspace{0.5em}

\noindent SI also demonstrates task adherence and is likewise able to adjust its style toward varying instructions (e.g., summarizing a text \textit{``in a tweet''}), but its outputs tend to be generic and lack the specificity seen in CRAFT's document-grounded summaries.
EI again introduces superficially complex phrasing, such as \textit{``Unraveling the Perplexing Enigma of Age-Related Pituitary [\dots]''}, or \textit{``In the intricately woven narrative of Afghanistan's rich and complex history [\dots]''}.

\section{Conclusion}
In this work, we introduced CRAFT (Corpus Retrieval and
Augmentation for Fine-Tuning), a framework for generating
task-specific synthetic datasets grounded in
text
corpora. CRAFT requires only a small set of human-curated few-shot examples to bootstrap the creation of large-scale training data by leveraging existing corpora and instruction-tuned language models. Our experiments across multiple tasks, including biology, medicine, and commonsense question-answering, as well as summarization, demonstrate that models fine-tuned on CRAFT-generated datasets can match or outperform strong baselines, including instruction-tuned models and those trained on human-curated datasets. Notably, CRAFT-based models showed better generalization capabilities on out-of-domain datasets compared to models trained on human-curated data and maintained robustness to variations in the quality of the initial few-shot examples.
Furthermore, while some fully synthetic methods such as Self-Instruct produce competitive results, CRAFT outperforms these approaches overall, offering a more scalable and reliable dataset generation framework.

While CRAFT shows promising results for most tasks, we also identified limitations in scaling performance for recipe generation, emphasizing the need for careful quality control and potential stopping criteria in future iterations.
Nevertheless, the overall success of CRAFT in producing high-quality synthetic datasets with minimal human effort opens up new possibilities for efficient and adaptable model fine-tuning across a wide range of domains and tasks.

\section*{Acknowledgements}
We thank the action editor, Tao Ge, and the anonymous reviewers at TACL for their helpful comments during the review process.
IZ and DE have been supported by the European Union’s Horizon 2020 research and innovation program under grant agreement No. 101135671 (TrustLLM).
AK and HS have been funded by Deutsche Forschungsgemeinschaft (DFG, German Research Foundation) project SCHU 2246/14-1.
IZ acknowledges the EuroHPC Joint Undertaking for awarding access to MareNostrum5, hosted at Barcelona Supercomputing Center (BSC), Spain, under proposal No. EHPC-DEV-2024D12-031.
IZ was also supported by a G-Research Travel Grant to present this work at the ELLIS Doctoral Symposium 2024 in Paris.

\bibliography{tacl2021}
\bibliographystyle{acl_natbib}

\iftaclpubformat

\onecolumn

\appendix

\section{Implementation Details}
\label{sec:implementation_details}

\subsection{Few-Shot Design}
\label{sec:appendix-few-shot-design}
\textbf{BioQA} few-shot texts were drawn from diverse sources, including textbooks~\citep{alberts2017molecular,MalmquistPrescott2022,WilkinBrainard2016,RyeEtAl2016}, Encyclopedia Britannica, and openly accessible materials.
\textbf{MedQA} examples were based on public content from health-related websites (e.g., NIH, NHS, FDA, Mayo Clinic, Cleveland Clinic).
\textbf{CSQA} few-shots were compiled from blogs, articles, and other topical online sources.
\textbf{Recipe generation} few-shots were sourced from blogs and public recipe sites, typically structured as an instruction or question with lists of ingredients and steps.
If recipes were in continuous prose, the authors added structured elements manually.
To ensure diverse retrieval, a wide range of dishes and vocabulary was included.
\textbf{Summarization} few-shots were built from texts across websites, blogs, magazines, and GitHub issues.
Each example included a full input text, a summarization instruction, and a corresponding summary.
Where summaries were not available, they were authored to match realistic format and content.

\subsection{Corpora}
To enable retrieving human-written documents for general-purpose as well as specialized domains, we include four large corpora.

\textbf{C4}~\citep{raffel2020exploring} provides a broad web-crawled dataset. We use a 305GB filtered subset of the original 750GB corpus, excluding not-safe-for-work content.
\textbf{Wikipedia} offers high-quality encyclopedic content. We use the English Wikipedia dump from January 2, 2024, processed with WikiExtractor~\citep{Wikiextractor2015} to extract clean text.
\textbf{Stack Exchange}~\citep{StackExchangeDataset} includes QA-format documents across 173 communities, combining title, body, and top-voted answer. It spans technical and non-technical domains.
\textbf{WikiHow}~\citep{koupaee2018wikihow} features instructional content in a step-by-step format, useful for generative tasks such as recipe generation or summarization.

After filtering documents to lengths between 200 and 25,000 characters, we retain 362M documents from C4, 10.5M from Wikipedia, 9.5M from Stack Exchange, and 190K from WikiHow. Combined, these 383M documents occupy 247GB when GZIP-compressed~\citep{deutsch1996gzip} and stored as 16-bit NumPy arrays~\citep{harris2020array}.

\subsection{Document Retrieval}
We use a two-step retrieval strategy to approximate global similarity search over 383M embedded documents, avoiding expensive full pairwise comparisons.

First, the embedding database is split into sequential shards of 350K documents. For each shard, cosine similarity to the few-shot samples is computed, and the top 5\% most similar documents are retained, reducing the candidate pool to 19M documents.
Secondly, we perform a second round of cosine similarity and standard top-$k$ retrieval on this reduced set to obtain the final document matches.

To avoid redundancy or topical overfitting in retrieval, we combine two strategies: (i) 50\% of samples are retrieved via top-$k$ similarity to each few-shot individually, and (ii) 50\% via similarity to the averaged embedding of all few-shots to capture latent task structure.

\subsection{Task Sample Synthesis}
Synthetic task samples are generated via in-context learning~\citep{brown2020language}, using three randomly sampled few-shot examples per prompt. These are interleaved with the instruction and a retrieved corpus document, following the template shown in Table~\ref{fig:corpus-to-task_sample}, step 3. Because each few-shot includes a long text, input prompts often exceed 10,000 tokens (up to 20,000), requiring long-context models.

We use Mistral 7B Instruct v0.2~\citep{jiang2023mistral} with vLLM~\citep{kwon2023efficient} for generation. Sampling is performed with temperature=0.7, top-$k$=40~\citep{fan2018hierarchical}, and top-$p$=0.9~\citep{holtzman2020curious}. Maximum output lengths are capped at 256 tokens for QA, 1280 for recipes, and 1536 for summarization, based on empirical tuning.

To enable quality control, all outputs are formatted as JSON with fixed keys. We discard any samples with malformed structure, missing fields, or insufficient length. For QA tasks, we validate that answer options are complete and contain a valid label. To reduce redundancy, we filter out samples with high similarity to few-shots or other outputs using fuzzy string matching~\citep{ranjith2022review} with a token set ratio $>0.85$. We recommend retrieving roughly twice the desired number of corpus samples to account for filtering. Task-specific filtering statistics are detailed in Appendix~\ref{sec:filtering_stats}.

\section{Additional Results: Varying the Number of Few-Shots}
\label{sec:additional-results-varying-few-shots}
\begin{figure*}[h]
    \includegraphics[width=\linewidth]{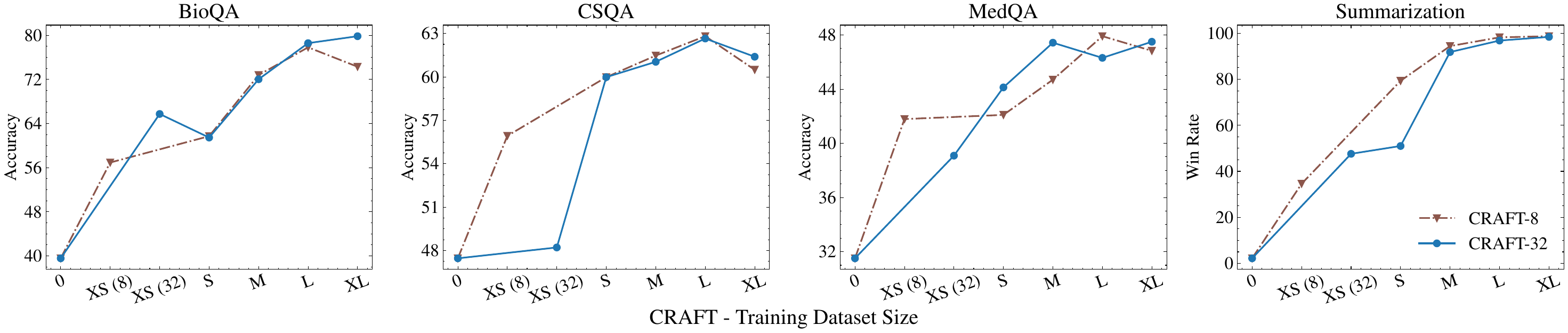}
    \caption{%
    Performance between CRAFT when initiated with 8 and 32 human-curated few-shot examples.
    }
\end{figure*}

\section{Full-Length CRAFT-Generated Examples}

\subsection{Recipe Generation}
\label{sec:appendix-recipe_samples}

\noindent\textbf{Instruction:} Prepare apple strudel using puff pastry and apples with sugar and raisins. \\
\textbf{Ingredients:} 5 medium apples, peeled, cored, and sliced; 1/2 cup granulated sugar; 1/2 cup raisins; 1 pound puff pastry, thawed; 1 egg, beaten; 1 tbsp water; powdered sugar, for dusting. \\
\textbf{Steps:} \textbf{Cook the apples:} In a saucepan, cook the sliced apples with sugar over medium heat until softened and caramelized. \textbf{Combine apples and raisins:} Transfer the cooked apples to a bowl and mix in the raisins. \textbf{Prepare the pastry:} Unroll the puff pastry on a lightly floured surface and place it on a baking sheet. \textbf{Add apple filling:} Place the apple filling in a line down the middle of the pastry, leaving a 2-inch border on all sides. \textbf{Brush the edges:} Brush the edges of the pastry with the beaten egg and water mixture. \textbf{Fold the pastry:} Fold the edges of the pastry over the apple filling, sealing it tightly. \textbf{Bake the strudel:} Brush the top with the egg mixture and bake at 375°F for 30–35 minutes, until golden brown. \textbf{Cool and serve:} Let cool slightly before dusting with powdered sugar and serving.

\bigskip

\noindent\textbf{Instruction:} Prepare a dish using items from your well-stocked pantry, freezer, and fridge. \\
\textbf{Ingredients:} 
Honey, Chipotle peppers in adobo sauce, Assorted olives, jalapenos, pepperoncinis etc., Flour (all-purpose, whole wheat, self-rising), Sugar, brown sugar, Powdered sugar, Baking powder, baking soda, extracts, yeast, Masa, Potatoes, onions, garlic, Dried beans, Stocks and broths (chicken, beef, vegetable), Jarred pesto and specialty relishes, chutneys, etc., Panko breadcrumbs, Kosher salt, black pepper, Worcestershire, Tabasco, olive oil, different vinegars, soy sauce, Longer lasting cheeses (cheddar, Parmesan, feta), Bacon, Corn tortillas, Beef, sausage (breakfast sausage, Italian sausage, chorizo), Dinner rolls, Pie crust, Frozen fruits (peaches, berries, cherries, etc.), Frozen vegetables (green beans, peas, lima beans, carrots, corn), Ready-made recipes, Chili, Homemade pumpkin puree

\noindent\textbf{Steps:} \textbf{Choose a recipe} or decide on a dish to make using the available ingredients. \textbf{Prepare and gather ingredients} from the pantry, freezer, and fridge. \textbf{Cook the dish} according to the recipe or your preferred method. \textbf{Serve and enjoy!}

\bigskip

\noindent\textbf{Instruction:} Create a salad or bowl by combining greens, proteins, vegetables, cheeses, fruits, and nuts with the specified dressing. \\
\textbf{Ingredients:}
Chopped cucumber, Tomato, Bell pepper, Red onion, Feta cheese, Kalamata olives, Chopped green kale, Shredded chicken, Black beans, Crumbled cojita, Diced avocado, Spicy crispy pepitas, Lemon-cumin vinaigrette, Chopped Napa cabbage, Rice noodles, Julienned red bell peppers, Scallions, Shredded carrots, Cilantro, Peanuts, Zingy miso ginger vinaigrette, Crispy noodles, Sweet corn, Diced bell pepper, Tomato, Avocado, Red onion, Cilantro, Toasted pine nuts, Blue cheese vinaigrette, Dried cherries, Mild feta cheese, Swiss chard, Peppery baby arugula, Toasted pecans, Champagne vinaigrette, Fresh baby lettuce, Sweet green peas, Crunchy bacon shards, Savory white mushroom sauté, Asiago cheese, Tart vinaigrette, Grilled Bosc pears, Bacon crumbles, Toasted pecan halves, Tangy Maytag blue cheese, Tender greens, Ruby port vinaigrette, Frisée greens, Poached egg, Bacon lardons, Warm shallot vinaigrette, Baby spinach, Sautéed mushrooms, Bacon, Hardboiled egg pieces, Slivered red onion, Honey balsamic vinaigrette, Red cabbage, Romaine, Tomato, Corn, Green and red onion, Seasoned beef and chorizo, Cheddar and jack cheese, Cumin honey-lime vinaigrette, Duck confit, Hoisin and hot sauce, Peanuts, Cabbage, Scallions, Carrots, Crisp jicama shells, Assorted fruits, Scoop of ice cream

\noindent\textbf{Steps:} \textbf{Chop or prepare} all vegetables, fruits, and proteins as specified in the text. \textbf{Combine the greens} in a large bowl or on individual plates. \textbf{Add proteins, vegetables, fruits, cheeses, and nuts} to the greens. \textbf{Drizzle with the specified dressing} or vinaigrette. \textbf{Toss gently} to evenly coat the salad or bowl. \textbf{Serve and enjoy!}

\subsection{Summarization}
\label{sec:appendix-summarization_samples}

\noindent\textbf{Instruction:} Summarize the text below, focusing on Bill Ackman's investment in Procter \& Gamble and his intentions for the company.

\noindent\textbf{Input Text:}
Bill Ackman, a well-known hedge fund manager, believes that Procter \& Gamble (P\&G) could be worth more than its current share price due to poor marketing and pricing strategies. Ackman holds around 1\% of the company's shares and has been actively advocating for changes. During a presentation, he mentioned that he spends 25\% of his time on his investment in P\&G. Ackman has been critical of the present management and its financial performance since disclosing his stake last year. It remains to be seen whether the company will welcome Ackman's involvement more enthusiastically than in the past. In the past, P\&G has been targeted by activists after an event-driven turn in the share price. Ackman has not specified who he would like to see as the new CEO but has mentioned that potential changes could lead to savings, marketing efficiencies, and operating leverage. He did not mention any particular candidates during his presentation.

\noindent\textbf{Generated Summary:}
Bill Ackman, a hedge fund manager, believes that Procter \& Gamble's (P\&G) stock could be worth more than its current price due to poor marketing and pricing strategies. He holds around 1\% of the company's shares and is actively pushing for changes. Ackman has been vocal about his dissatisfaction with the current management and has spent significant time advocating for improvements. He has not specified who he would like to see as the new CEO but has mentioned that potential changes could lead to savings, marketing efficiencies, and operating leverage.

\section{Filtering Statistics}
\label{sec:filtering_stats}

\begin{table}[htpb]
\centering
\begin{tabular}{lcccc}
\toprule
\textbf{Filter Criteria} & \textbf{S} & \textbf{M} & \textbf{L} & \textbf{XL} \\
\midrule
Retrieved Corpus Samples & 240 & 1,200 & 12,000 & 60,000 \\
\midrule
Exact duplicates & 25 & 37 & 819 & 8,551 \\
Excessive length & 2 & 14 & 266 & 1,632 \\
Format errors & 10 & 40 & 466 & 2,174 \\
Similarity to few-shots & 0 & 1 & 22 & 45 \\
Similarity to other task samples & 9 & 117 & 1,469 & 5,961 \\
\midrule
Available synthetic task samples & 194 & 991 & 8,958 & 41,637 \\
\bottomrule
\end{tabular}
\caption{%
BioQA corpus and task sample filtering process.
Corpus samples are turned into task samples after filtering for excessive length.
}
\end{table}

\begin{table}[htpb]
\centering
\begin{tabular}{lcccc}
\toprule
\textbf{Filter Criteria} & \textbf{S} & \textbf{M} & \textbf{L} & \textbf{XL} \\
\midrule
Retrieved Corpus Samples & 240 & 1,200 & 12,000 & 60,000 \\
\midrule
Exact duplicates & 24 & 30 & 165 & 1,348 \\
Excessive length & 2 & 8 & 64 & 307 \\
Format errors & 5 & 30 & 364 & 1,879 \\
Similarity to few-shots & 11 & 19 & 141 & 410 \\
Similarity to other task samples & 14 & 129 & 2,655 & 17,749 \\
\midrule
Available synthetic task samples & 184 & 984 & 8,611 & 38,307 \\
\bottomrule
\end{tabular}
\caption{%
CSQA corpus and task sample filtering process.
Corpus samples are turned into task samples after filtering for excessive length.
}
\end{table}

\begin{table}[htpb]
\centering
\begin{tabular}{lcccc}
\toprule
\textbf{Filter Criteria} & \textbf{S} & \textbf{M} & \textbf{L} & \textbf{XL} \\
\midrule
Retrieved Corpus Samples & 240 & 1,200 & 12,000 & 60,000 \\
\midrule
Exact duplicates & 24 & 24 & 50 & 773 \\
Excessive length & 1 & 10 & 141 & 890 \\
Format errors & 15 & 36 & 540 & 2,911 \\
Similarity to few-shots & 0 & 10 & 55 & 204 \\
Similarity to other task samples & 3 & 40 & 813 & 5,221 \\
\midrule
Available synthetic task samples & 197 & 1,080 & 10,401 & 50,001 \\
\bottomrule
\end{tabular}
\caption{%
MedQA corpus and task sample filtering process.
Corpus samples are turned into task samples after filtering for excessive length.
}
\end{table}

\begin{table}[htpb]
\centering
\begin{tabular}{lcccc}
\toprule
\textbf{Filter Criteria} & \textbf{S} & \textbf{M} & \textbf{L} & \textbf{XL} \\
\midrule
Retrieved Corpus Samples & 240 & 1,200 & 12,000 & 60,000 \\
\midrule
Exact duplicates & 24 & 24 & 28 & 620 \\
Excessive length & 1 & 1 & 20 & 54 \\
Format errors & 87 & 417 & 4,035 & 18,684 \\
Similarity to few-shots & 6 & 18 & 111 & 389 \\
Similarity to other task samples & 0 & 7 & 473 & 3,711 \\
\midrule
Available synthetic task samples & 122 & 733 & 7,333 & 36,542 \\
\bottomrule
\end{tabular}
\caption{%
RecipeGen corpus and task sample filtering process.
Corpus samples are turned into task samples after filtering for excessive length.
}
\end{table}

\begin{table}[htpb]
\centering
\begin{tabular}{lcccc}
\toprule
\textbf{Filter Criteria} & \textbf{S} & \textbf{M} & \textbf{L} & \textbf{XL} \\
\midrule
Retrieved Corpus Samples & 240 & 1,200 & 12,000 & 60,000 \\
\midrule
Exact duplicates & 24 & 24 & 19 & 101 \\
Excessive length & 34 & 189 & 1,793 & 8,964 \\
Format errors & 55 & 336 & 3,119 & 14,803 \\
Similarity to few-shots & 21 & 28 & 99 & 379 \\
Similarity to other task samples & 1 & 1 & 32 & 394 \\
\midrule
Available synthetic task samples & 105 & 622 & 6,938 & 35,359 \\
\bottomrule
\end{tabular}
\caption{%
Summarization corpus and task sample filtering process.
Corpus samples are turned into task samples after filtering for excessive length.
}
\end{table}

\clearpage
\section{Dataset Statistics}
\label{sec:dataset_stats}

\begin{table}[htpb]
\centering
\begin{tabular*}{\textwidth}{l@{\extracolsep{\fill}}cccccccc}
\toprule
\multirow{2}{*}{\textbf{Dataset}} & \multirow{2}{*}{\textbf{Size}} & \multicolumn{2}{c}{\textbf{Corpus Samples}} & \multicolumn{2}{c}{\textbf{TS Instruction}} & \multicolumn{2}{c}{\textbf{TS Output}} \\
\cmidrule(lr){3-4} \cmidrule(lr){5-6} \cmidrule(lr){7-8}
& & Mean & Median & Mean & Median & Mean & Median \\
\midrule
\multirow{5}{*}{BioQA} & XS & 1,109 & 1,088 & 93 & 91 & \multirow{5}{*}{1} & \multirow{5}{*}{1} \\
& S & 1,786 & 1,170 & 83 & 77 & & \\
& M & 1,858 & 1,093 & 76 & 64 & & \\
& L & 2,033 & 1,038 & 80 & 69 & & \\
& XL & 2,122 & 972 & 86 & 77 & & \\
\midrule
\multirow{5}{*}{CSQA} & XS & 1,496 & 1,444 & 25 & 26 & \multirow{5}{*}{1} & \multirow{5}{*}{1} \\
& S & 1,265 & 851 & 25 & 25 & & \\
& M & 1,399 & 884 & 26 & 25 & & \\
& L & 1,324 & 864 & 26 & 25 & & \\
& XL & 1,300 & 848 & 27 & 26 & & \\
\midrule
\multirow{5}{*}{MedQA} & XS & 1,755 & 1,815 & 117 & 118 & \multirow{5}{*}{1} & \multirow{5}{*}{1} \\
& S & 1,612 & 1,203 & 85 & 77 & & \\
& M & 1,577 & 1,053 & 79 & 67 & & \\
& L & 1,599 & 1,013 & 78 & 68 & & \\
& XL & 1,691 & 1,001 & 81 & 71 & & \\
\midrule
\multirow{5}{*}{RecipeGen} & XS & 1,277 & 1,223 & 16 & 16 & 593 & 528 \\
& S & 1,168 & 823 & 20 & 19 & 433 & 363 \\
& M & 1,107 & 807 & 24 & 22 & 369 & 326 \\
& L & 1,058 & 786 & 24 & 23 & 355 & 319 \\
& XL & 1,005 & 754 & 24 & 23 & 345 & 316 \\
\midrule
\multirow{5}{*}{Summarization} & XS & 1,595 & 734 & 1,019 & 690 & 82 & 61 \\
& S & 1,442 & 829 & 612 & 442 & 107 & 92 \\
& M & 1,440 & 852 & 471 & 366 & 116 & 106 \\
& L & 1,396 & 880 & 432 & 358 & 122 & 110 \\
& XL & 1,369 & 882 & 433 & 355 & 117 & 107 \\
\bottomrule
\end{tabular*}
\caption{%
Dataset Statistics.
TS is short for task sample.
For summarization, the instruction includes the model-generated long but cleaned text augmentation from a corpus sample that will subsequently be summarized.
}
\end{table}

\fi

\end{document}